\documentclass[10pt,twocolumn,letterpaper]{article}

\usepackage{iccv}
\usepackage{times}
\usepackage{epsfig}
\usepackage{graphicx}
\usepackage{amsmath}
\usepackage{amssymb}
\usepackage{multicol}
\usepackage{multirow}
\usepackage{arydshln}
\usepackage{pifont}
\usepackage{booktabs}
\usepackage{float}
\usepackage{caption}
\usepackage{subcaption}
\usepackage{varwidth}

\usepackage[T1]{fontenc}
\usepackage[font=small,labelfont=bf,tableposition=top]{caption}

\DeclareCaptionLabelFormat{andtable}{#1~#2  \&  \tablename~\thetable}

\usepackage[accsupp]{axessibility}  

\newcommand{\ignore}[1]{}

\usepackage[pagebackref=true,breaklinks=true,letterpaper=true,colorlinks,bookmarks=false]{hyperref}

\iccvfinalcopy 

\def\iccvPaperID{8312} 

\begin{document}

\title{Mean Shift for Self-Supervised Learning}

\author{\fontsize{11}{11} \selectfont Soroush Abbasi Koohpayegani$^{1,}\footnotemark[1]\quad$ 
 Ajinkya Tejankar$^{1,}\thanks{Equal contribution}\quad$ Hamed Pirsiavash$^{1,2}\quad$   \\
 \\
 
\fontsize{11}{11} \selectfont $^{1}$ University of Maryland, Baltimore County $\quad$  $^{2}$University of California, Davis\\


}

\maketitle

\begin{abstract}

Most recent self-supervised learning (SSL) algorithms learn features by contrasting between instances of images or by clustering the images and then contrasting between the image clusters. We introduce a simple mean-shift algorithm that learns representations by grouping images together without contrasting between them or adopting much of prior on the structure or number of the clusters. We simply ``shift'' the embedding of each image to be close to the ``mean'' of the neighbors of its augmentation. Since the closest neighbor is always another augmentation of the same image, our model will be identical to BYOL when using only one nearest neighbor instead of $5$ used in our experiments. Our model achieves $72.4\%$ on ImageNet linear evaluation with ResNet50 at $200$ epochs outperforming BYOL. Also, our method outperforms the SOTA by a large margin when using weak augmentations only, facilitating adoption of SSL for other modalities.
Our code is available here: \textcolor{magenta}{\href{https://github.com/UMBCvision/MSF}{https://github.com/UMBCvision/MSF}}

\end{abstract}

\section{Introduction}

Most current visual recognition algorithms are supervised, meaning that they learn from large scale annotated images or videos. However, in many applications, the annotation process may be expensive, biased, ambiguous, or involve privacy concerns. Self-supervised learning (SSL) algorithms aim to learn rich representations from unlabeled images or videos. Such learned representations can be used along with small annotated data to provide an accurate visual recognition model. We are interested in developing better SSL models using unlabeled images.


Some recent SSL models learn by contrasting between instances of images. They pull different augmentations of an image instance together while pushing them away from other image instances \cite{he2020momentum,chen2020simple}. Some other SSL methods cluster the unlabeled images to a set of clusters with the hope that each cluster will contain semantically similar images. Then, a model that predicts those clusters learns rich representations similar to supervised learning with labels \cite{caron2018deep,asano2020self,caron2020unsupervised}.

These clustering methods also can be considered as contrastive learning since they contrast between different clusters of images. For instance, the SoftMax layer in deep clustering method \cite{caron2018deep} encourages an image to be assigned to the correct single cluster and not the other clusters.

Also, most clustering algorithms have strong priors on the overall structure of the clusters. For instance, deep clustering (k-means) using Euclidean distance encourages spherical cluster shapes which we believe is unnecessary for the purpose of SSL methods. 



Recently, BYOL \cite{grill2020bootstrap} showed that it is possible to learn rich representations without contrasting between image instances. BYOL \cite{grill2020bootstrap} works by simply pulling the two views of an image closer without any contrast with other images. The better performance of BYOL \cite{grill2020bootstrap} compared to MoCo hints that contrasting with other images may be a limiting constraint. For instance, in MoCo \cite{he2020momentum}, since the negative images are sampled randomly, they may be from the same category as the query, resulting in degraded representations. \cite{tejankar2020isd} tries to resolve this issue by not treating all negatives equally negative.




Inspired by mean shift clustering, we generalize BYOL to a simple yet effective SSL method where a data point is pulled closer to not only its other augmentations but also the nearest neighbors (NNs) of its augmentation. Unlike DeepCluster \cite{caron2018deep}, SwAV \cite{caron2020unsupervised}, and SeLA \cite{asano2020self} that use explicit, mutually exclusive cluster assignment, our method uses mean-shift algorithm that groups similar images together locally without explicit cluster assignment. Moreover, unlike k-means clustering, mean-shift does not have strong priors on the shape, size, or number of the clusters. This makes mean-shift suitable for SSL where such priors are not known. Compared to MoCo \cite{he2020momentum}, SimCLR \cite{chen2020simple}, SwAV \cite{caron2020unsupervised} and few others, our method never contrasts between different images or even cluster centers.

\begin{figure*}[ht!]
\begin{center}
\end{center}
   \includegraphics[width=1.0\linewidth]{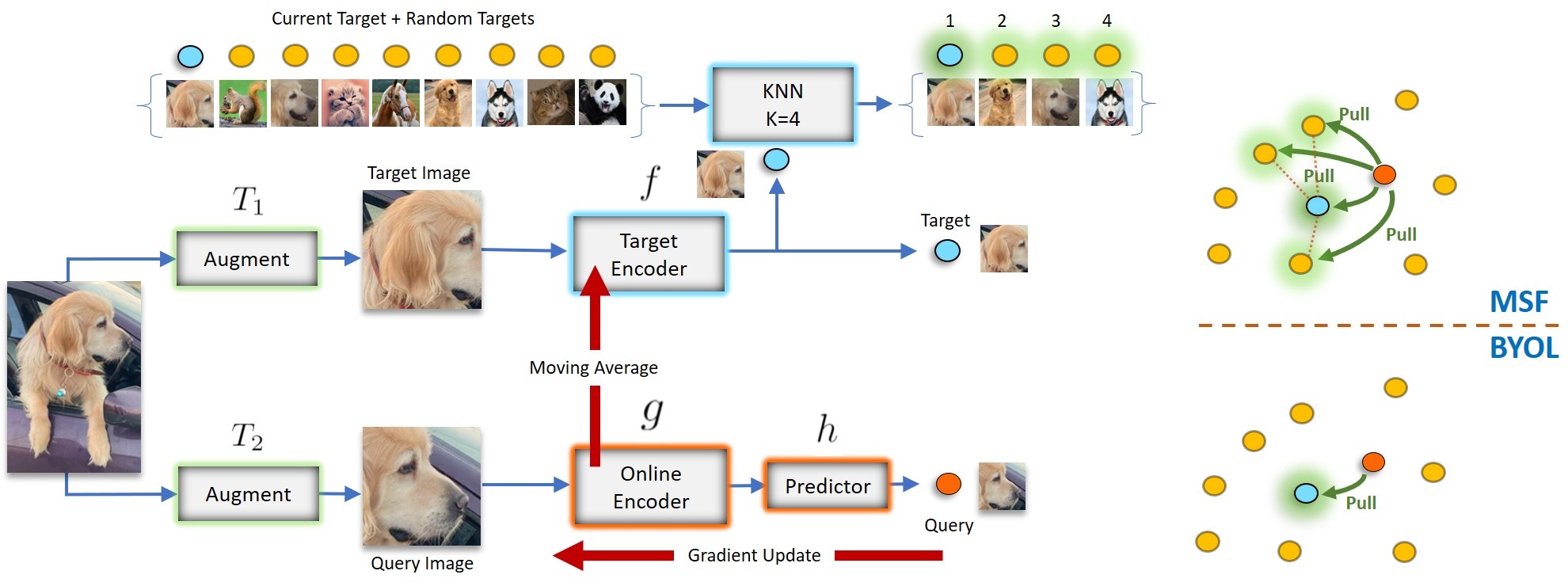}
   \caption{{\bf MSF method:} Similar to BYOL, we maintain two encoders (``target'' and ``online''). The online encoder is updated with gradient descent while the target encoder is the moving average of the online encoder. We augment an image twice and feed to both encoders. We add the target embedding to the memory bank and look for its nearest neighbors in the memory bank. Obviously target embedding itself will be the first nearest neighbor. We want to shift the embedding of the input image towards the mean of its nearest neighbors, so we minimize the summation of those distances. Note that our method using only one nearest neighbor is identical to BYOL which pulls different augmentations together without grouping different instances of images. To our knowledge, our method is the first in grouping different instances of images without contrasting between image instances or clusters.}
\label{fig:Method}
\end{figure*}

Since we need a large set of embeddings to search for nearest neighbors, we adopt the memory bank idea \cite{he2020momentum} to maintain a random set of embeddings. Also, since the model is evolving over time in the learning process, the old elements in the memory bank will not be valid, so we adopt the momentum idea from \cite{he2020momentum} to maintain two encoders (``target'' and ``online'') instead of only one. The online model is updated by the loss and the target model is updated as a moving average of the online model. We feed two different augmentations of an image to these two encoders, then we push the online embedding of the image to be close to the average of nearest neighbors of the target encoding of the image in the target embedding space. Hence, similar to most recent SSL methods, our method also uses the inductive bias that the augmentation should not move the embedding much. 


Our experiments show that our method outperforms state-of-the-art methods on various settings. For instance, when trained on unlabeled ImageNet for 200 epochs, it achieves $72.4\%$ linear ImageNet accuracy which is better than BYOL at 200 epochs. 

Most recent SSL methods use strong augmentations to improve the accuracy, leading to ``augmentation engineering'' to improve SSL. However, in many applications, e.g., medical domain, designing such augmentations is not easy and needs extensive domain knowledge. Hence, designing SSL methods that do not rely heavily on large variations of augmentations is interesting. We show that when using only weak augmentations, our method (MSF w/w) outperforms BYOL by a large margin. We hypothesize that NNs act as a proxy for strong augmentations of the query image, so there is no need for engineering strong augmentations. 





\begin{figure*}
\begin{center}
\end{center}
   \includegraphics[width=1.0\linewidth]{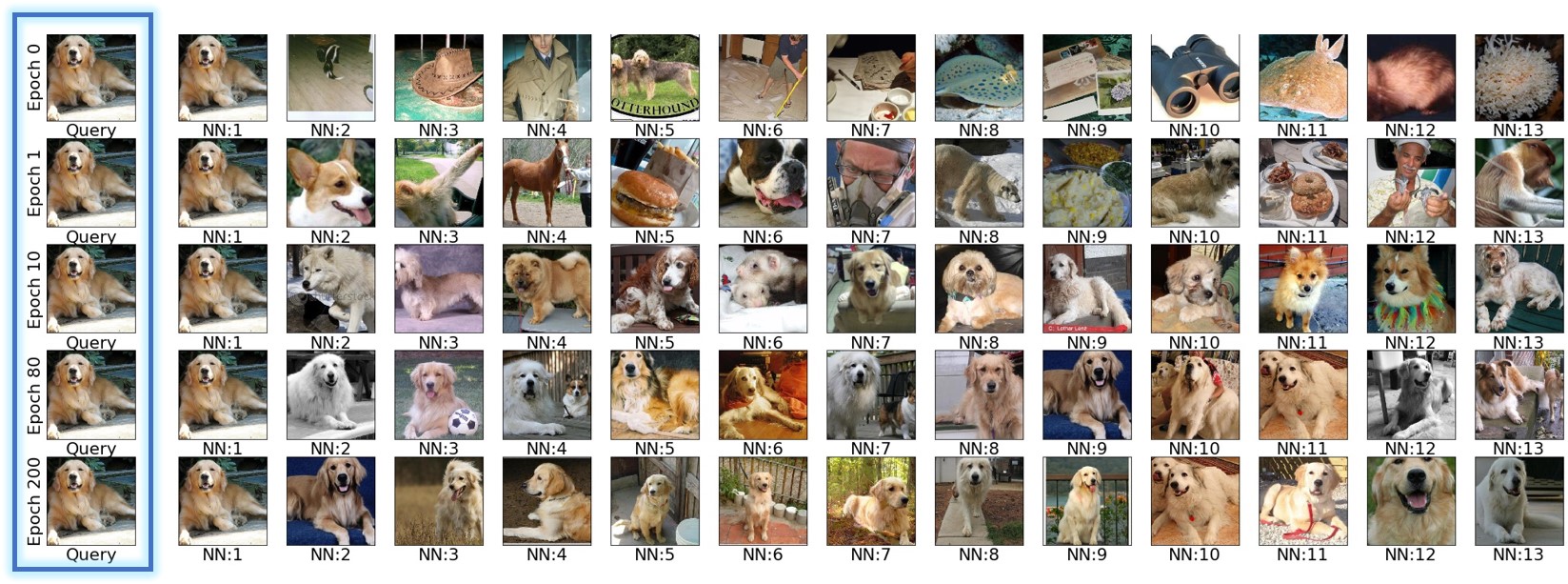}
   \caption{{\bf Nearest neighbors (NN) of the model at each epoch:} For a random query images, we show how the nearest neighbors evolve at the learning time. Initially, NNs are not semantically quite related, but are close in low-level features. The accuracy of 1-NN classifier in the initialization is $1.5\%$ which is $15$ times larger than random chance ($0.1\%$). This little signal is bootstrapped in our learning method and results in NNs of the late epochs which are mostly semantically related to the query image. More visualizations can be found in the appendix.}
\label{fig:vis_NN}
\end{figure*}

\section{Method}
We are interested in mean-shift clustering, so at each iteration we want to encourage the model to shift the embedding of an image to be closer to the average of its nearest neighbors on a large random set of samples. 


Following the notation of BYOL \cite{grill2020bootstrap}, we assume a target encoder $f$ and an online encoder $g$. Both encoders have the same backbone architecture followed by a projection layer and are initialized equally. The online encoder $g$ is followed by an additional prediction layer $h$ on top of it. The online encoder $g$ and the prediction layer $h$ are updated by back-propagating the loss while the target encoder $f$ is updated by momentum update to be a running average of the online encoder $g$. Since nearest neighbor needs a large pool of examples, we maintain a first-in-first-out (FIFO) memory bank \cite{he2020momentum} that includes recent embeddings from the slowly evolving target encoder $f$. 


Given an unlabeled image $x$, we augment it randomly twice to get $T_1(x)$ and $T_2(x)$. We feed them to encoders and then normalize them with $\ell_2$ norm to get $u=\frac{f(T_1(x))}{||f(T_1(x))||_2}$ and $v=\frac{h(g(T_2(x)))}{||h(g(T_2(x)))||_2}$. We first add $u$ to the memory bank and then, find the $k$ nearest neighbors of $u$ in the memory bank to get a set of embeddings $\{z_j\}_{j=1}^k$. Note that this set includes $u$ itself. Since we know it is another augmentation of the same input image, it should be a good target for $v$. Finally, we minimize the following loss:
$$L=\frac{1}{k}\sum_{j=1}^k dist(v, z_j)$$

\noindent where, $dist(.,.)$ is the distance metric between two embeddings. We use MSE loss ($dist(a,b)=||a-b||_2^2$) as the distance in our experiments. Minimizing this loss is equivalent to maximizing Cosine similarity as the vectors are already $\ell_2$ normalized.
The final loss is the summation of the above loss for all input images. 

Ideally, we can average the set of nearest neighbors to come up with a single target embedding, but since averaging is dependent on the choice of loss function, we simply minimize the summation of distances. Note that for Euclidean distance, both methods result in identical gradients. 


Since $u$ itself is included in our NN search, it will be always the best nearest neighbor. Hence, our method with $k=1$ will be identical to BYOL which minimizes $||v-u||_2^2$ for each image without using a memory bank. 

Moreover, in the initial stages of the learning, $v$ may be far from $u$ and the other $k-1$ nearest neighbors may be semantically different from the query image. Since those wrong neighbors are still close to $u$, the loss will still pull $v$ closer to the neighborhood of $u$ (another augmentation of $v$).
In later stages of learning, when the representation is more mature, the other $k-1$ neighbors will be semantically related and will contribute to learning since $u$ and $v$ are already closer to each other. Table \ref{tab:imagenet}-right and Figure \ref{fig:vis_NN} show that the representation improves as the learning progresses.



{\bf Strength of augmentation:} In most exemplar-based SSL methods, augmentation plays an important role since the main supervision signal is that the augmentation should not change the embedding much. Hence, recent methods, e.g., MoCo v2, SimCLR, and BYOL, use strong augmentations. We believe such aggressive augmentations on the target embeddings $u$ may add randomness to the learning process as some of those augmentation do not look natural, so the nearest neighbors will not be semantically close to the query image. Hence, we use weaker augmentations for the target model to make $u$ and $z$ less noisy while still using strong augmentations for the online model. We refer to this as the weak/strong (``w/s'') variation. This is inspired by \cite{sohn2020fixmatch} which uses weak augmentations in semi-supervised learning. This variation results in almost one point improvement over the regular variation where both encoders use strong augmentations. As shown in Fig. \ref{fig:purtiy} (right), the nearest neighbors are more pure in the ``weak only'' setting which is consistent with our above intuition. Our experiments show that BYOL also benefits from w/s augmentation to some extent. This is probably due to more robust target encoding. Finally, we explore a weak/weak ``w/w'' variation where both views are augmented with a weak augmentation.
 


\section{Experiments}

We report the results of our self-supervised learning and transfer learning in this section. We use PyTorch library for all experiments.



\subsection{Self-supervised Learning}

 \textbf{Mean Shift (MSF):} 
We use $0.99$ for the the momentum of the target encoder, top-$k$ = 5, and $1.024$M for the memory bank size (which is roughly the same as the size of ImageNet dataset). Our ablation study shows that a memory bank of $128$K does not degrade the results. We observe that the added computational cost of NN search is small compared to the overall forward and backward passes. We find that MSF with 128K memory bank size and 512 dimensions for the embedding, uses less than 0.5GB of extra GPU memory for the memory bank and less than $1\%$ of extra computation for finding 5 NNs (see Table \ref{tab:cost}).

\begin{table}[t]
    \centering
    \scalebox{1.0}{
    \begin{tabular}{lccccc}
        \toprule      
        Mem. & kNN & kNN & NN & 20-NN & Transfer \\
        Size &   Time & \small{GFLOPS} & & & Mean \\
        \midrule
        1M & 6.78\% & 1.05 & \textbf{62.0} & 64.9 & 75.5 \\
        128K & 0.72\% & 0.13 & \textbf{62.0} & \textbf{65.2} & \textbf{76.3} \\
        \bottomrule
    \end{tabular}
    }
    \caption{{\bf Additional computational cost of finding NNs:} Forwarding through each ResNet50 encoder needs 4.14 GFLOPS, so finding NNs adds a small cost. Note that any method like MoCo that uses a memory bank needs this additional cost.}  
    \label{tab:cost}
\end{table}



 \textbf{BYOL-asym (baseline):} Training SSL methods for more than $200$ epochs is not easy due to resource constraints. For instance, training BYOL with ResNet50 for $200$ epochs takes roughly $7$ days on four RTX 2080-Ti GPUs. Thus, for a fair comparison, we re-implement BYOL in our own framework and call it BYOL-asym. We note and justify the major differences between BYOL-asym and BYOL here. First, we use an asymmetric loss. The original BYOL paper \cite{grill2020bootstrap} uses symmetric loss which passes each view of the image through both encoders. As a result, the gradient is calculated over $2 \times B$ instances where $B$ is the batch size, so each epoch needs twice computation compared to asymmetric loss. Hence, 200 epochs of BYOL-asym should be compared with $100$ epochs of regular BYOL.
Second, we use a small batch size of 256 instead of 4096. \cite{grill2020bootstrap} shows that BYOL works well even with the batch size of 256. Third, we use SGD optimizer instead of LARS. Despite these differences, our implementation works reasonable well compared to reported results in prior work. Our MSF uses the same setup for fairness.



 \textbf{Augmentation:} In all of our experiments, ``strong'' augmentation refers to the augmentation in MoCo v2 \cite{chen2020mocov2}. The strong augmentation involves the following stochastic operations: grayscale, color jitter, horizontal flip, and Gaussian blur. The ``weak'' augmentation is simply a random crop of size $224\times224$ with area ratio between $0.2$ and $1.0$ followed by random horizontal flipping with probability $0.5$. \textbf{MSF w/s} refers to our ``weak/strong'' variation where the target encoder view is augmented with the weak augmentation and online encoder view is augmented with the strong augmentation. \textbf{MSF w/w} refers to our ``weak/weak'' variation where both teacher and student views use weak augmentation. \textbf{BYOL-asym} and \textbf{MSF} use the standard SSL practice of augmenting both views with the strong augmentation. 


 \textbf{Architecture:} We generally follow \cite{grill2020bootstrap} for the architectures of both BYOL-asym and MSF. We use the ResNet50 \cite{he2016deep} model as backbone in all our experiments. A projection layer (2 layer MLP) is added on top of the backbone. The first layer expands the feature channels from 2048 to 4096. It is followed by BatchNorm and ReLU layers. The final linear layer reduces the feature channels from 4096 to 512. The prediction layer architecture is the same as projection layer except its first layer expands the channels from 512 to 4096. 
 After the pre-training step, online encoder's backbone is evaluated by removing the projection and prediction layers. 

 \textbf{Training:} For both BYOL-asym and MSF, we use the SGD (lr=0.05, momentum=0.9, and weight decay=1e-4) optimizer and train for 200 epochs. Learning rate uses cosine scheduler.



\subsection{Evaluation on ImageNet}


 \textbf{Evaluation on full ImageNet.} We evaluate the representations of the pre-trained model by training linear and nearest neighbor (NN) classifiers. We use the code provided by \cite{abbasi2020compress} for training both classifiers. A single linear layer is trained on top of a frozen backbone. The features from the backbone are normalized to have unit $\ell_2$ norm and then scaled and shifted to have zero mean and unit variance for each dimension. The linear layer is trained with SGD (lr=0.01, epochs=40, batch size=256, weight decay=1e-4, and momentum=0.9). Learning rate is multiplied by 0.1 at 15 and 30 epochs. We use standard supervised ImageNet augmentations \cite{official_pytorch_imagenet_train} during training. For nearest neighbor, we pre-process train and val ImageNet sets with center crop (size 256) augmentation and calculate $\ell_2$ normalized embeddings by forwarding throught the backbone. We report Top-1 accuracies on ImageNet val set for linear, 1-NN, and 20-NN classifiers in Table \ref{tab:imagenet}.

\begin{table*}[!ht]
    \centering
    \begin{tabular}{cc}

    \begin{tabular}{lccccccc}
        \toprule
        Method & Ref. & Batch & Epochs & Sym. Loss& Top-1 & NN & 20-NN \\
        & & Size & & 2x FLOPS & Linear & & \\ 
        \midrule
        Supervised & \cite{official_pytorch_models} & 256 & 100 & - & 76.2 & 71.4 & 74.8 \\
        \midrule
        Random-init & - & - & - & - & 5.1 & 1.5 & 2.0  \\
        SeLa-v2 \cite{asano2020self} & \cite{caron2020unsupervised} & 4096 & 400 & \ding{51} & 67.2 & - & - \\
        SimCLR\cite{chen2020simple} & \cite{chen2020simple} & 4096 & 1000 & \ding{51} & 69.3 & - & - \\
        SwAV \cite{caron2020unsupervised} & \cite{caron2020unsupervised} & 4096 & 400 & \ding{51} & 70.1 & - & - \\
        DeepCluster-v2 \cite{caron2018deep} & \cite{caron2020unsupervised} & 4096 & 400 & \ding{51} & 70.2 & - & - \\
        SimSiam \cite{chen2020exploring} & \cite{chen2020exploring} & 256 & 400 & \ding{51} & 70.8 & - & - \\
        MoCo v2 \cite{he2020momentum} & \cite{chen2020exploring} &  256 & 400  & \ding{51} & 71.0 & - & - \\
        MoCo v2 \cite{he2020momentum} & \cite{chen2020mocov2} & 256 & 800 & \ding{55} & 71.1 & 57.3 & 61.0 \\
        CompRess \textsuperscript{$\dagger$} \cite{abbasi2020compress} & \cite{abbasi2020compress} & 256 & 1K+130 & \ding{55} & 71.9 & 63.3& 66.8\\
        InvP & \cite{wang2020invp} & 256 & 800 & \ding{55} & 71.3 & - & - \\
        BYOL \cite{grill2020bootstrap} & \cite{grill2020bootstrap} & 256 & 300 & \ding{51} & 71.8 & - & - \\
        BYOL \cite{grill2020bootstrap} & \cite{grill2020bootstrap} & 4096 & 1000 & \ding{51} & 74.3 & 62.8 & 66.9 \\
        SwAV \textsuperscript{$\ddagger$} \cite{caron2020unsupervised} & \cite{caron2020unsupervised} & 4096 & 800 & \ding{51} & \textbf{75.3} & - & - \\
        \midrule
        SimCLR\cite{chen2020simple} & \cite{chen2020exploring} & 4096 & 200 & \ding{51} & 68.3 & - & - \\
        SwAV \cite{caron2020unsupervised} & \cite{chen2020exploring} & 4096 & 200 & \ding{51} & 69.1 & - & - \\
        MoCo v2 \cite{he2020momentum} & \cite{chen2020exploring} &  256 & 200  & \ding{51} & 69.9 & - & - \\
        SimSiam \cite{chen2020exploring} & \cite{chen2020exploring} & 256 & 200 & \ding{51} & 70.0 & - & - \\
        BYOL \cite{grill2020bootstrap} & \cite{chen2020exploring} & 4096 & 200 & \ding{51} & \textbf{70.6} & - & - \\
        \midrule
        MoCo v2 \cite{he2020momentum} & \cite{chen2020mocov2} & 256 & 200 & \ding{55} & 67.5 & 50.9 & 54.3 \\
        CO2 \cite{wei2020co2} & \cite{wei2020co2} & 256 & 200 & \ding{55} & 68.0 & - & - \\
        BYOL-asym \cite{grill2020bootstrap} & - & 256 & 200 & \ding{55} & 69.3 & 55.0 & 59.2 \\
        ISD \cite{tejankar2020isd} & \cite{tejankar2020isd} & 256 & 200 & \ding{55} & 69.8 & 59.2 & 62.0 \\
        MSF & - & 256 & 200 & \ding{55} & 71.4 & 60.6 & 64.0 \\
        MSF w/s & - & 256 & 200 & \ding{55} & \bf{72.4} & \textbf{62.0} & 64.9 \\
        MSF w/s (128K) & - & 256 & 200 & \ding{55} & 72.1 & \textbf{62.0} & \textbf{65.2} \\
        \midrule
        \midrule
        SimCLR w/w \cite{chen2020simple} & \cite{grill2020bootstrap} & 4096 & 300 & \ding{51} & 40.2 & - & - \\
        BYOL w/w \cite{grill2020bootstrap} & \cite{grill2020bootstrap} & 4096 & 300 & \ding{51} & 60.1 & - & - \\
        MSF w/w & - & 256 & 200 & \ding{55} & \textbf{66.3} & 54.6 & 57.4 \\
        \bottomrule
        \end{tabular}
        &
        \includegraphics[height=0.27\textwidth,angle=-90,trim=29cm 0 0 -5cm]{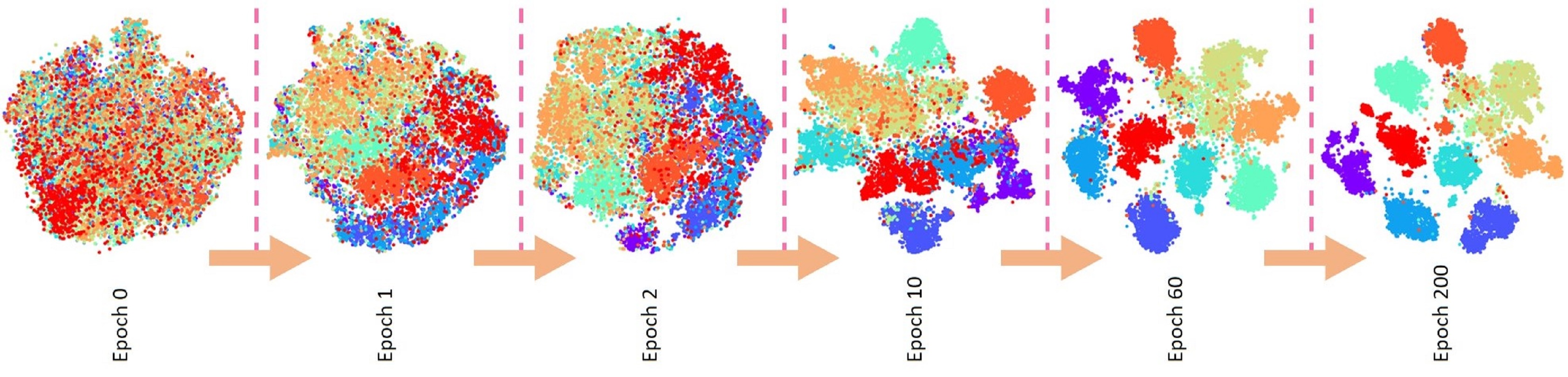}

    \end{tabular}
    \vspace{.1in}
    \caption{\textbf{Left: Evaluation on full ImageNet: } We compare our model on the full ImageNet linear and nearest neighbor benchmarks using ResNet50. We find that given similar computational budget, our models are better than other state-of-the-art methods. Our w/s variation works slightly better than the regular MSF. Interestingly, when using weak augmentations only, our method (MSF w/w) outperforms BYOL and SimCLR with a large margin. We believe this is important in some applications,e g.g. medical domain, where augmentation engineering is not easy. Note that methods with symmetric loss are not directly comparable with the other ones as they need to feed each image twice though each encoder. This results in twice the computation for each mini-batch. One may argue that a non-symmetric BYOL with 200 epochs should be compared with symmetric BYOL with 100 epochs only as they use similar amount of computation. Note that symmetric MoCo v2 with 400 epochs is almost the same as asymmetric MoCo v2 with 800 epochs ($71.0$ vs. $71.1$). Note that the accuracy of Random-init for ResNet50 is much lower than AlexNet (14.1\% on conv5 layer from \cite{noroozi2018boosting}). $\dagger$: CompRess is not directly comparable as it uses ResNet50 distilled from a larger SSL teacher model (SimCLR-ResNet50x4). $\ddagger$: SwAV is not comparable as it uses multiple crops together. \textbf{Right: Epochwise t-SNE for MSF: } We visualize the $\ell_2$ normalized features for 10 random ImageNet classes at certain epochs of MSF training. We find that over the period of training, semantic clusters are formed in the feature space. 
    }
    \vspace{-,1in}
\label{tab:imagenet}
\end{table*}




 \textbf{Evaluation on smaller ImageNet:} Similar to \cite{henaff2019data,chen2020simple,grill2020bootstrap,abbasi2020compress}, we evaluate the pre-trained models on the task of classification with limited ImageNet labels. The training details are the same as above except the training dataset sizes are reduced to 1\% and 10\% of the train set of ImageNet \cite{ILSVRC15}. The results are reported in Table \ref{tab:lim_sup}.

\begin{table}[th!]
    \begin{center}
    \scalebox{0.87}{
    \begin{tabular}{lcccccc}
        \toprule
        \multirow{2}{*}{Method} & Fine- & \multirow{2}{*}{Epochs} & \multicolumn{2}{c}{Top-1} & \multicolumn{2}{c}{Top-5} \\
        & tuned & & 1\% & 10\% & 1\% & 10\% \\
        \midrule
        Supervised & \ding{51} & & 25.4 & 56.4 & 48.4 & 80.4 \\
        PIRL \cite{misra2019self} & \ding{51} & 800 & - & - & 57.2 & 83.8  \\
        CO2 \cite{wei2020co2} & \ding{51} & 200 & - & - & 71.0 & 85.7 \\
        SimCLR \cite{chen2020simple} & \ding{51} & 1000 & 48.3 & 65.6 & 75.5 & 87.8 \\
        InvP \cite{wang2020invp} & \ding{51} & 800 & - & - & 78.2 & 88.7 \\
        BYOL \cite{grill2020bootstrap} & \ding{51} & 1000 & 53.2 & 68.8 & 78.4 & 89.0 \\
        $\text{SwAV}^{\dagger}$ \cite{caron2020unsupervised} & \ding{51} & 800 & \textbf{53.9} & \textbf{70.2} & \textbf{78.5} & \textbf{89.9} \\
        \midrule
        MoCo v2 \cite{chen2020mocov2} & \ding{55} & 800 & 51.5 & 63.6 & 77.6 & 86.1 \\
        $\text{BYOL}^{\ddag}$ \cite{grill2020bootstrap} & \ding{55} & 1000 & 55.7 & \textbf{68.6} & 80.0 & \textbf{88.6} \\
        CompRess* \cite{abbasi2020compress} & \ding{55} & 1K+130 & \textbf{59.7} & 67.0 & \textbf{82.3} & 87.5 \\
        \midrule
        MoCo v2 \cite{chen2020mocov2} & \ding{55} & 200 & 43.6 & 58.4 & 71.2 & 82.9 \\
        BYOL-asym  & \ding{55} & 200 & 47.9 & 61.3 & 74.6 & 84.7 \\
        ISD \cite{tejankar2020isd} & \ding{55} & 200 & 53.4 & 63.0 & 78.8 & 85.9 \\
        MSF & \ding{55} & 200 & 53.5 & 65.2 & 78.1 & 86.4 \\
        MSF w/s & \ding{55} & 200 & \textbf{55.5} & \textbf{66.5} & \textbf{79.9} & \textbf{87.6} \\
        \bottomrule
    \end{tabular}
    }
    \end{center}
    \caption{\textbf{Evaluation on small labeled ImageNet: } We compare our model on the ImageNet 1\% and 10\% linear evaluation benchmarks for ResNet50. The column ``Fine-tuned'' refers to whether the full network was fine-tuned or a single linear layer was trained. Given similar computational budgets, both of our models are better than other state-of-the-art methods. We evaluate BYOL and MoCo v2 with our evaluation framework and interestingly, realize that BYOL performs better in linear evaluation compared to finetuning the whole network on the 1\% split. We report these numbers for fair comparison. * is ResNet50 compressed from SimCLR-R50x4. $\dagger$ uses a different augmentation strategy than others. $\ddag$ is our evaluation with the official weights \cite{byolweights}. 
    }
    \label{tab:lim_sup}
    \vspace{-.1in}
\end{table}

\subsection{Evaluation on Transfer Learning}

\begin{table*}[th!]
    \begin{center}
    \scalebox{0.88}{
    \begin{tabular}{l|c|c|c|c|c|c|c|c|c|c|c|c|c}
    \toprule
    Method & Ref. & Epochs & Food & CIFAR & CIFAR & SUN & Cars & Aircraft & DTD & Pets & Caltech & Flowers & Mean \\
    & & & 101 & 10 & 100 & 397 & 196 &  &  &  & 101 & 102 & \\
    \midrule
    Supervised & \cite{grill2020bootstrap} &  & 72.3 & 93.6 & 78.3 & 61.9 & 66.7 & 61.0 & 74.9 & 91.5 & 94.5 & 94.7 & 78.9 \\
    \midrule
    SimCLR \cite{chen2020simple} & \cite{grill2020bootstrap} & 1000 & 72.8 & 90.5 & 74.4 & 60.6 & 49.3 & 49.8 & 75.7 & 84.6 & 89.3 & 92.6 & 74.0 \\
    MoCo v2 \cite{chen2020mocov2} & - & 800 & 72.5 & 92.2 & 74.6 & 59.6 & 50.5 & 53.2 & 74.4 & 84.6 & 90.0 & 90.5 & 74.2 \\
    BYOL \cite{grill2020bootstrap} & \cite{grill2020bootstrap} & 1000  & 75.3 & 91.3 & \textbf{78.4} & \textbf{62.2} & \textbf{67.8} & 60.6 & 75.5 & \textbf{90.4} & \textbf{94.2} & \textbf{96.1} & \textbf{79.2} \\
    BYOL \cite{grill2020bootstrap} & rep. & 1000 & \textbf{75.4} & \textbf{92.7} & 78.1 & 62.1 & 67.1 & \textbf{62.0} & \textbf{76.8} & 89.8 & 92.2 & 95.5 & \textbf{79.2} \\
    \midrule
    
    BYOL-asym \cite{grill2020bootstrap} & - & 200 & 70.2 & 91.5 & 74.2 & 59.0 & 54.0 & 52.1 & \textbf{73.4} & 86.2 & 90.4 & 92.1 & 74.3 \\
    MoCo v2 \cite{chen2020mocov2} & - & 200 & 70.4 & 91.0 & 73.5 & 57.5 & 47.7 & 51.2 & 73.9 & 81.3 & 88.7 & 91.1 & 72.6 \\
    
    MSF & - & 200 & 70.7 & 92.0 & 76.1 & 59.0 & \textbf{60.9} & 53.5 & 72.1 & 89.2 & 92.1 & 92.4 & 75.8 \\
    MSF-w/s & - & 200 & 71.2 & 92.6 & \textbf{76.3} & 59.2 & 55.6 & 53.7 & 73.2 & 88.7 & \textbf{92.7} & 92.0 & 75.5 \\
    MSF-w/s (128K) & - & 200 & \textbf{72.3} & \textbf{92.7} & \textbf{76.3} & \textbf{60.2} & 59.4 & \textbf{56.3} & 71.7 & \textbf{89.8} & 90.9 & \textbf{93.7} & \textbf{76.3} \\
    \bottomrule
    \end{tabular}
    }
    \end{center}
    \caption{\textbf{Linear layer transfer learning evaluation: } We compare various SSL methods on transfer tasks by training linear layers. Under similar computational budgets, we show that our models are consistently better or on par with other state-of-the-art methods. Only a single linear layer is trained on top of features. No train time augmentations are used. ``rep.'' means we have reproduced the results using our evaluation framework for better comparison. 
    }
    \label{tab:transfer_linear}
\end{table*}

\textbf{Linear classification:} Following the procedure outlined in \cite{chen2020simple,grill2020bootstrap}, we evaluate the self-supervised pre-trained models for linear classification task on following datasets: Food101 \cite{food101}, SUN397 \cite{sun397}, CIFAR10 \cite{cifar}, CIFAR100 \cite{cifar}, Cars \cite{carsdataset}, Aircraft \cite{aircraft}, Flowers \cite{flowers}, Pets \cite{pets}, Caltech-101 \cite{caltech101} and DTD \cite{dtd}. The appendix includes more details on the datasets and training. The results are reported in Table \ref{tab:transfer_linear}. To verify our implementation, we evaluate the official 1000-epoch BYOL weights provided in \cite{byolweights} and compare with the results from \cite{grill2020bootstrap} in Table \ref{tab:transfer_linear}.



 \textbf{Object detection:} Following the procedure outlined in \cite{he2020momentum}, we use Faster-RCNN \cite{ren2015faster} for the task of object detection on PASCAL-VOC \cite{everingham2010pascal}. We use the code provided at \cite{official_moco_code} with default parameters. All the weights are finetuned on the \texttt{trainval07+12} set and evaluated on the \texttt{test07} set. We report an average over 5 runs in Table \ref{tab:detection}.

\begin{table}[th!]
    \begin{center}
    \scalebox{1.0}{
        \begin{tabular}{lccccc}
            \toprule
            Method & Ref. & Epochs & $\text{AP}_{50}$ & AP & $\text{AP}_{75}$ \\
            \midrule
            Sup. IN & \cite{chen2020exploring} & - & 81.3 & 53.5 & 58.8 \\
            Scratch & \cite{chen2020exploring} & - & 60.2 & 33.8 & 33.1 \\
            \midrule
            \multicolumn{6}{l}{\textit{Symmetric loss. 2x FLOPS}} \\
            SimCLR & \cite{chen2020exploring} & 200 & 81.8 & 55.5 & 61.4 \\
            MoCo v2 & \cite{chen2020exploring} & 200 & 82.3 & 57.0 & 63.3 \\
            BYOL & \cite{chen2020exploring} & 200 & 81.4 & 55.3 & 61.1 \\
            SwAV & \cite{chen2020exploring} & 200 & 81.5 & 55.4 & 61.4 \\
            SimSiam & \cite{chen2020exploring} & 200 & 82.4 & 57.0 & 63.7 \\
            \midrule
            \multicolumn{6}{l}{\textit{Asymmetric loss.}} \\
            MoCo v2 & \cite{chen2020mocov2} & 800 & 82.5 & 57.4 & 64.0 \\
            InvP  & \cite{wang2020invp} & 800 & 81.8 & 56.2 & 61.5 \\
            MoCo v2 & \cite{chen2020mocov2} & 200 & 82.4 & 57.0 & 63.6 \\
            CO2 & \cite{wei2020co2} & 200 & \textbf{82.7} & \textbf{57.2} & \textbf{64.1} \\
            BYOL-asym  & - & 200 & 81.9 & 56.8 & 63.5 \\
            MSF & - & 200 & 82.2 & 56.7 & 63.4 \\
            MSF w/s & - & 200 & 82.2 & 56.6 & 63.1 \\
            \bottomrule
        \end{tabular}
    }
    \end{center}
    \caption{\textbf{Transfer learning to PASCAL VOC object detection: } We compare our models on the transfer task of object detction. We find that given a similar computational budget, our method is better than BYOL. The models are trained on the VOC \texttt{trainval07+12} set and evaluated on the \texttt{test07} set. We report average over 5 runs. 
    }
    \label{tab:detection}
    \vspace{-.15in}
\end{table}

\subsection{Ablation study.}

Here, we study the effect of MSF hyperparameters and design choices like augmentation strategies, top-\textit{k}, and memory bank size. We use ResNet50 and train it with ImageNet. In all experiments, we use the default MSF w/s variant and only vary the parameter of interest.


{\bf Same view of an instance to both encoders:}
One may argue that the mean-shift grouping and using different views of the same instance for different encoders are orthogonal ideas, and mean-shift alone might work. We did an experiment by feeding the same augmented view to both encoders ($T1=T2$) and realized that the model does not learn. It collapses in the first epoch. Hence, we believe using different views is still an important inductive bias.



\begin{figure*}
     \centering
     \begin{subfigure}[b]{0.32\textwidth}
         \centering
         \includegraphics[width=\textwidth]{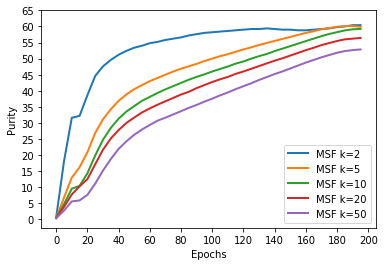}
     \end{subfigure}
     \hfill
     \begin{subfigure}[b]{0.32\textwidth}
         \centering
         \includegraphics[width=\textwidth]{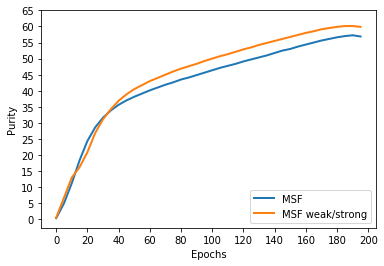}
     \end{subfigure}
     \hfill
     \begin{subfigure}[b]{0.31\textwidth}
         \centering
         \includegraphics[width=\textwidth]{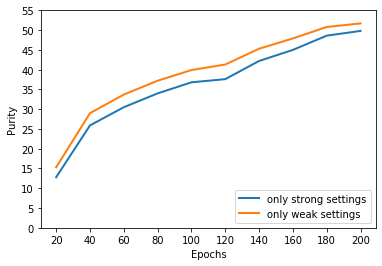}
     \end{subfigure}
     \hfill
        \caption{\textbf{Comparison of purity: } Ideally, we want all $k$ nearest neighbors to be from the same category as the input image, so we calculate the percentage of correct neighbors for each input, average it over all images, and call it purity. We exclude the first NN which is identical to $u$, so the accuracy is over $k-1$ neighbors. We find this metric to be very handy in evaluating the model at the training time since it comes almost for free. {\bf (left)} shows the purity for different $k$ values with respect to epoch number while {\bf(middle)} compares the purity of ``w/s'' variation with the regular ``s/s'' variation for $k=5$. Purity is higher for ``w/s'' variation which is consistent with our intuition. In {\bf (right)}, at each epoch of our MSF w/s (top-$k=10$) model, we calculate purity for the target encoder using either only weak (orange) or only strong (blue) augmentations. We see that strong setting has lower purity. This suggests that the stronger augmentation makes the nearest neighbors more noisy. This is aligned with our intuition for using weak augmentation for the target model in w/s variation.}
        
         
        \label{fig:purtiy}
\end{figure*}

\textbf{Effect of \textit{k} in top-\textit{k}:} This section shows the effect of sampling different top-\textit{k} nearest neighbors. We use \textit{k} values from set $\{2, 5, 10, 20, 50\}$. We use $k=5$ for main experiments, but $k=10$ improves NN by $0.5$ point.
Note that setting $k=1$, makes MSF identical to BYOL. Results are in the Table \ref{tab:abl_msf_topk}. Additionally, we plot the purity for each experiment in Figure \ref{fig:purtiy}. Purity for a single query is the percentage of the samples $2$ to $k$ in the top-\textit{k} nearest neighbors (excluding $u$ itself) which have the same class as the query. Final purity is calculated by averaging the purities of all samples. One may study the effect of increasing $k$ gradually during iterations as a future extension.

\begin{figure}[th!]
\begin{center}
\end{center}
  \includegraphics[width=1.0\linewidth]{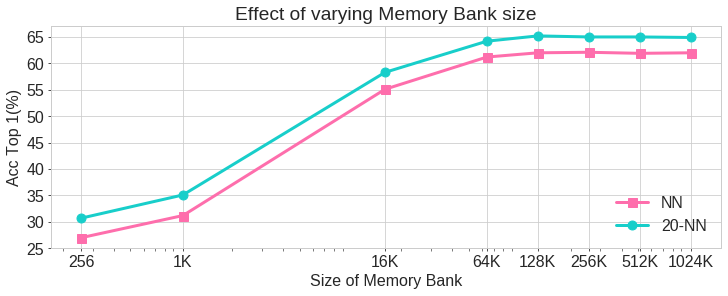}
  \vspace{-.25in}
  \caption{\textbf{Memory Bank Size:} On ImageNet, we do not see an improvement in increasing the memory bank size beyond 128K which needs only 0.5GB of GPU memory.}
\label{fig:abl_msf_membank_size}
\end{figure}

\begin{table}[h!]
    \begin{center}
    \scalebox{1.0}{
    \begin{tabular}{ccccccc}
        \toprule
        & \textit{k}=1 & \textit{k}=2 & \textit{k}=5 & \textit{k}=10 & \textit{k}=20 & \textit{k}=50 \\
        \midrule
        NN & 55.8 & 61.0 & 62.0 & 62.5 & 62.0 &  61.5 \\
        20-NN & 59.1 & 64.2 & 64.9 & 65.7 & 65.4 & 64.9 \\
        \bottomrule
    \end{tabular}
    }
    \end{center}
    \caption{\textbf{Effect of \textit{k} in top-\textit{k}:} Our study shows that MSF is not very sensitive to $k$. While $k=10$ performs the best, we report the main results for $k=5$.
Note that setting $k=1$ makes our method identical to BYOL-asym.}
    \label{tab:abl_msf_topk}
    \vspace{-.3in}
\end{table}


 \textbf{Size of memory bank:} 
CompRess \cite{abbasi2020compress} shows that a large memory bank is important to accurately capture the neighborhood of a random sample in the embedding space. Thus, we vary the size of the memory bank from 256 to 1M to evaluate if larger memory bank can help with more accurate nearest neighbors. Results are in Figure \ref{fig:abl_msf_membank_size}. Although our main experiment uses 1M sized memory bank, we find that 128K works equally well. Note that the size of memory bank also depends on the training dataset size. 

 \textbf{Comparison of different augmentation strategies:} Table \ref{tab:byol_weak_strong} shows results for BYOL and MSF with different augmentation strategies. Comparing ``s/s'' variants with ``w/s'', we find that BYOL receives a very small boost from the ``w/s'' variant while MSF improves consistently by $\approx1$ point on all three benchmarks. We believe this is due to better purity of the nearest neighbors while training (also shown in Fig. \ref{fig:purtiy} (right)). Further, we observe that MSF w/w is significantly better as compared to BYOL w/w. This can be attributed to the nearest neighbors serving as a proxy for strong augmentation.

\section{Related Work}

 \textbf{Self-supervised learning: } The aim of self-supervised learning is to learn representations directly from the data without using any manual data annotation. Specifically, a pretext task is designed based on the inherent structure in the data and a model is trained to solve it. Various pretext tasks have been designed that exploit different structural cues in the data. Here, we focus on following pretext tasks for images: treating each data point as a single class to perform instance level classification \cite{dosovitskiy2014discriminative}, predicting the relative location of patches \cite{doersch2015unsupervised,noroozi2016unsupervised}, filling up a missing patch in an image \cite{pathak2016context}, predicting a colored image from its grayscale version \cite{zhang2016colorful,zhang2017split}, counting objects in an image \cite{noroozi2017representation}, predicting the rotation of an image \cite{gidaris2018unsupervised}, and predicting the pseudo-labels obtained from clustering \cite{caron2018deep,asano2020self}. Note that designing the pretext task or the augmentations itself is still manual and needs domain knowledge.
 

 \textbf{Instance discrimination:} Recently, the task of instance discrimination \cite{dosovitskiy2014discriminative} has shown great promise. The basic idea is to treat each image as single class. This is also referred as contrastive learning where positive samples (augmented views of the same instance) are pushed close and away from the negative samples (all other instances). While \cite{dosovitskiy2014discriminative} took a parametric approach for this classification, \cite{wu2018unsupervised} took a non-parametric approach. The non-parametric approach has seen broad adoption with great results \cite{he2020momentum,chen2020simple,misra2019self,zhuang2019local,tian2019cmc,caron2020unsupervised}. Two important components of these methods are: memory bank (source of negative samples) and augmentation (constructing positive samples). A simple but effective technique using a momentum encoder to populate the memory bank is proposed in \cite{he2020momentum}. A rigorous study of the impact of different augmentations and hyperparameters is conducted in \cite{chen2020simple}. Improved augmentation strategies are proposed in \cite{caron2020unsupervised,chen2020simple,tian2020makes}.  Instance discrimination can also be viewed from an information theoretic perspective as the task of maximizing the information between different views of a single image \cite{hjelm2018learning,oord2018representation,bachman2019learning}.

 \textbf{Consistency regularization:} Although negative pairs were thought to be central in preventing the collapse of representations for instance discrimination, \cite{grill2020bootstrap} proposed a method that does not collapse despite not using any negatives. The objective in \cite{grill2020bootstrap,chen2020exploring} simply pulls augmented views of the same image close without any contrast with negative samples. This is also referred to as consistency regularization in the semi-supervised learning framework \cite{tarvainen2017mean}. Inspired by \cite{grill2020bootstrap}, we propose a more general form of it where the positives can also come from the close neighborhood of a sample grouping similar images together.  


 \textbf{Clustering methods: } Another class of methods based on clustering have shown promise. The basic idea is to alternate between clustering and learning the representations \cite{xie2016dec,yangCVPR2016joint}. This approach was first scaled to large scale pre-training in \cite{doersch2015unsupervised}. A big concern in these methods is to prevent the collapse of all representations into a single cluster. To that end, an optimal transport based formulation of clustering is proposed in \cite{asano2020self}. An online clustering algorithm based on the formulation of \cite{asano2020self} is proposed in \cite{caron2020unsupervised}.

 \textbf{Clustering and instance discrimination: } Clustering methods can be seen as the generalizing the instance discrimination framework. Only the views from the same sample can be positives in instance discrimination \cite{dosovitskiy2014discriminative,wu2018unsupervised}, but all the members (and their views) of a cluster are positives in clustering based methods \cite{bautista2016clique,caron2020unsupervised}. A more flexible middle ground is where the set of positives are based on the local neighborhood of a sample: top-\textit{k} nearest neighbors of a sample in \cite{huang2019unsupervised}, nearest neighbors of a sample that are also the members of the same cluster in \cite{zhuang2019local}, and top-\textit{k} graph distance based neighbors in \cite{wang2020invp}. Our method shares the motivation behind these works: embeddings should be locally clustered around high density regions. We also use top-\textit{k} nearest neighbors as positives \cite{huang2019unsupervised,wang2020invp}, but our method is fundamentally different as there is no notion of negatives in our method. Intuitively, we enforce a more simpler and flexible constraint: move each sample closer to its nearest neighbors in each iteration. This idea is inspired from Mean-Shift Clustering \cite{cheng1995mean,comaniciu2002mean} where the cluster assignment for each sample is iteratively updated to be the mean of its nearest neighbors. In contrast to the \textit{k}-means clustering, Mean-Shift does not make strong assumptions about the shape of clusters.

\begin{table}[t]
    \begin{center}
    \scalebox{1.0}{
        \begin{tabular}{lcccc}
        \toprule
        Method & Aug. & Top-1 & NN & 20-NN \\
        \midrule
        BYOL-asym & s/s & 69.3 & 55.0 & 59.2 \\
        BYOL-asym & w/s & 69.5 & 55.8 & 59.1 \\
        BYOL \textsuperscript{$\dagger$} \cite{grill2020bootstrap} & w/w & 60.1 & - & - \\
        MSF & s/s & 71.4 & 60.6 & 64.0 \\
        MSF & w/s & 72.4 & 62.0 & 64.9 \\
        MSF & w/w & 66.3 & 54.6 & 57.4 \\
        \bottomrule
        \end{tabular}
    }
    \vspace{0.1in}
    \caption{{\bf Comparison augmentations strategies:} In s/s, both views are strongly augmented while in w/w they are weakly augmented. w/s refers to weak augmentation for target view and strong augmentation for the online view. w/s improves our method more compared to BYOL. This may be due to more pure nearest neighbors. In w/w setting, MSF is significantly better than BYOL as the nearest neighbors can be a good substitute for strong augmentation. $\dagger$: uses 4096 batch size, 300 epochs, and symmetric loss.}
    \label{tab:byol_weak_strong}
    \end{center}
    \vspace{-.4in}
\end{table}


\section{Conclusion}
We introduce a simple but effective SSL method based on grouping similar images together in an online fashion. We simply shift the embedding of an image towards the mean of its nearest neighbors. MSF with $k=1$ is identical to BYOL so MSF can be seen as a generalized form of BYOL. Our extensive experiments show that MSF performs better or on-par compared to state-of-the-art SSL methods on various tasks including ImageNet linear evaluation. 

\noindent {\bf Acknowledgment:} 
This material is based upon work partially supported by the United States Air Force under Contract No. FA8750‐19‐C‐0098, funding from SAP SE, and also NSF grant numbers 1845216 and 1920079. Any opinions, findings, and conclusions or recommendations expressed in this material are those of the authors and do not necessarily reflect the views of the United States Air Force, DARPA, or other funding agencies.

{\small
\bibliographystyle{ieee_fullname}
\bibliography{egbib}

\begin{thebibliography}{10}\itemsep=-1pt

\bibitem{byolweights}
Code and weights for byol.
\newblock \url{https://github.com/deepmind/deepmind-research/tree/master/byol}.

\bibitem{official_pytorch_imagenet_train}
Official pytorch supervised imagenet training code.
\newblock
  \url{https://github.com/pytorch/examples/blob/master/imagenet/main.py}.

\bibitem{official_moco_code}
Pytorch implementation of moco: https://arxiv.org/abs/1911.05722.
\newblock \url{https://github.com/facebookresearch/moco}.

\bibitem{official_pytorch_models}
Torchvision models.
\newblock \url{https://pytorch.org/docs/stable/torchvision/models.html}.

\bibitem{abbasi2020compress}
Soroush Abbasi~Koohpayegani, Ajinkya Tejankar, and Hamed Pirsiavash.
\newblock Compress: Self-supervised learning by compressing representations.
\newblock {\em Advances in Neural Information Processing Systems}, 33, 2020.

\bibitem{bachman2019learning}
Philip Bachman, R~Devon Hjelm, and William Buchwalter.
\newblock Learning representations by maximizing mutual information across
  views.
\newblock In {\em Advances in Neural Information Processing Systems}, pages
  15535--15545, 2019.

\bibitem{bautista2016clique}
Miguel~A Bautista, Artsiom Sanakoyeu, Ekaterina Tikhoncheva, and Bjorn Ommer.
\newblock Cliquecnn: Deep unsupervised exemplar learning.
\newblock In {\em Advances in Neural Information Processing Systems},
  volume~29. Curran Associates, Inc., 2016.

\bibitem{food101}
Lukas Bossard, Matthieu Guillaumin, and Luc Van~Gool.
\newblock Food-101 -- mining discriminative components with random forests.
\newblock In {\em European Conference on Computer Vision}, 2014.

\bibitem{caron2018deep}
Mathilde Caron, Piotr Bojanowski, Armand Joulin, and Matthijs Douze.
\newblock Deep clustering for unsupervised learning of visual features.
\newblock In {\em Proceedings of the European Conference on Computer Vision
  (ECCV)}, pages 132--149, 2018.

\bibitem{caron2020unsupervised}
Mathilde Caron, Ishan Misra, Julien Mairal, Priya Goyal, Piotr Bojanowski, and
  Armand Joulin.
\newblock Unsupervised learning of visual features by contrasting cluster
  assignments.
\newblock In {\em Advances in Neural Information Processing Systems}, pages
  9912--9924. Curran Associates, Inc., 2020.

\bibitem{chen2020simple}
Ting Chen, Simon Kornblith, Mohammad Norouzi, and Geoffrey Hinton.
\newblock A simple framework for contrastive learning of visual
  representations.
\newblock {\em arXiv preprint arXiv:2002.05709}, 2020.

\bibitem{chen2020mocov2}
Xinlei Chen, Haoqi Fan, Ross Girshick, and Kaiming He.
\newblock Improved baselines with momentum contrastive learning.
\newblock {\em arXiv preprint arXiv:2003.04297}, 2020.

\bibitem{chen2020exploring}
Xinlei Chen and Kaiming He.
\newblock Exploring simple siamese representation learning.
\newblock {\em arXiv preprint arXiv:2011.10566}, 2020.

\bibitem{cheng1995mean}
Yizong Cheng.
\newblock Mean shift, mode seeking, and clustering.
\newblock {\em IEEE transactions on pattern analysis and machine intelligence},
  17(8):790--799, 1995.

\bibitem{dtd}
Mircea Cimpoi, Subhransu Maji, Iasonas Kokkinos, Sammy Mohamed, and Andrea
  Vedaldi.
\newblock Describing textures in the wild.
\newblock In {\em Computer Vision and Pattern Recognition}, 2014.

\bibitem{comaniciu2002mean}
Dorin Comaniciu and Peter Meer.
\newblock Mean shift: A robust approach toward feature space analysis.
\newblock {\em IEEE Transactions on pattern analysis and machine intelligence},
  24(5):603--619, 2002.

\bibitem{doersch2015unsupervised}
Carl Doersch, Abhinav Gupta, and Alexei~A Efros.
\newblock Unsupervised visual representation learning by context prediction.
\newblock In {\em Proceedings of the IEEE International Conference on Computer
  Vision}, pages 1422--1430, 2015.

\bibitem{dosovitskiy2014discriminative}
Alexey Dosovitskiy, Jost~Tobias Springenberg, Martin Riedmiller, and Thomas
  Brox.
\newblock Discriminative unsupervised feature learning with convolutional
  neural networks.
\newblock In {\em Advances in neural information processing systems}, pages
  766--774, 2014.

\bibitem{everingham2010pascal}
Mark Everingham, Luc Van~Gool, Christopher~KI Williams, John Winn, and Andrew
  Zisserman.
\newblock The pascal visual object classes (voc) challenge.
\newblock {\em International journal of computer vision}, 88(2):303--338, 2010.

\bibitem{caltech101}
Li Fei-Fei, Rob Fergus, and Pietro Perona.
\newblock Learning generative visual models from few training examples: An
  incremental bayesian approach tested on 101 object categories.
\newblock {\em Computer Vision and Pattern Recognition Workshop}, 2004.

\bibitem{gidaris2018unsupervised}
Spyros Gidaris, Praveer Singh, and Nikos Komodakis.
\newblock Unsupervised representation learning by predicting image rotations.
\newblock In {\em International Conference on Learning Representations}, 2018.

\bibitem{grill2020bootstrap}
Jean-Bastien Grill, Florian Strub, Florent Altch{\'e}, Corentin Tallec,
  Pierre~H Richemond, Elena Buchatskaya, Carl Doersch, Bernardo~Avila Pires,
  Zhaohan~Daniel Guo, Mohammad~Gheshlaghi Azar, et~al.
\newblock Bootstrap your own latent: A new approach to self-supervised
  learning.
\newblock {\em arXiv preprint arXiv:2006.07733}, 2020.

\bibitem{he2020momentum}
Kaiming He, Haoqi Fan, Yuxin Wu, Saining Xie, and Ross Girshick.
\newblock Momentum contrast for unsupervised visual representation learning.
\newblock In {\em Proceedings of the IEEE/CVF Conference on Computer Vision and
  Pattern Recognition}, pages 9729--9738, 2020.

\bibitem{he2016deep}
Kaiming He, Xiangyu Zhang, Shaoqing Ren, and Jian Sun.
\newblock Deep residual learning for image recognition.
\newblock In {\em Proceedings of the IEEE conference on computer vision and
  pattern recognition}, pages 770--778, 2016.

\bibitem{henaff2019data}
Olivier~J H{\'e}naff, Aravind Srinivas, Jeffrey De~Fauw, Ali Razavi, Carl
  Doersch, SM Eslami, and Aaron van~den Oord.
\newblock Data-efficient image recognition with contrastive predictive coding.
\newblock {\em arXiv preprint arXiv:1905.09272}, 2019.

\bibitem{hjelm2018learning}
R~Devon Hjelm, Alex Fedorov, Samuel Lavoie-Marchildon, Karan Grewal, Phil
  Bachman, Adam Trischler, and Yoshua Bengio.
\newblock Learning deep representations by mutual information estimation and
  maximization.
\newblock In {\em International Conference on Learning Representations}, 2019.

\bibitem{huang2019unsupervised}
Jiabo Huang, Qi Dong, Shaogang Gong, and Xiatian Zhu.
\newblock Unsupervised deep learning by neighbourhood discovery.
\newblock In {\em International Conference on Machine Learning}, pages
  2849--2858. PMLR, 2019.

\bibitem{carsdataset}
Jonathan Krause, Michael Stark, Jia Deng, and Li Fei-Fei.
\newblock {3D} object representations for fine-grained categorization.
\newblock In {\em Workshop on 3D Representation and Recognition}, Sydney,
  Australia, 2013.

\bibitem{cifar}
Alex Krizhevsky.
\newblock Learning multiple layers of features from tiny images.
\newblock Technical report, University of Toronto, 2009.

\bibitem{aircraft}
Subhransu Maji, Esa Rahtu, Juho Kannala, Matthew~B. Blaschko, and Andrea
  Vedaldi.
\newblock Fine-grained visual classification of aircraft.
\newblock {\em arXiv preprint arXiv:1306.5151}, 2013.

\bibitem{misra2019self}
Ishan Misra and Laurens van~der Maaten.
\newblock Self-supervised learning of pretext-invariant representations.
\newblock {\em arXiv preprint arXiv:1912.01991}, 2019.

\bibitem{flowers}
Maria-Elena Nilsback and Andrew Zisserman.
\newblock Automated flower classification over a large number of classes.
\newblock In {\em Indian Conference on Computer Vision, Graphics and Image
  Processing}, 2008.

\bibitem{noroozi2016unsupervised}
Mehdi Noroozi and Paolo Favaro.
\newblock Unsupervised learning of visual representations by solving jigsaw
  puzzles.
\newblock In {\em European Conference on Computer Vision}, pages 69--84.
  Springer, 2016.

\bibitem{noroozi2017representation}
Mehdi Noroozi, Hamed Pirsiavash, and Paolo Favaro.
\newblock Representation learning by learning to count.
\newblock In {\em Proceedings of the IEEE International Conference on Computer
  Vision}, pages 5898--5906, 2017.

\bibitem{noroozi2018boosting}
Mehdi Noroozi, Ananth Vinjimoor, Paolo Favaro, and Hamed Pirsiavash.
\newblock Boosting self-supervised learning via knowledge transfer.
\newblock In {\em Proceedings of the IEEE Conference on Computer Vision and
  Pattern Recognition}, pages 9359--9367, 2018.

\bibitem{pets}
O.~M. Parkhi, A. Vedaldi, A. Zisserman, and C.~V. Jawahar.
\newblock Cats and dogs.
\newblock In {\em Computer Vision and Pattern Recognition}, 2012.

\bibitem{pathak2016context}
Deepak Pathak, Philipp Krahenbuhl, Jeff Donahue, Trevor Darrell, and Alexei~A
  Efros.
\newblock Context encoders: Feature learning by inpainting.
\newblock In {\em Proceedings of the IEEE conference on computer vision and
  pattern recognition}, pages 2536--2544, 2016.

\bibitem{ren2015faster}
Shaoqing Ren, Kaiming He, Ross Girshick, and Jian Sun.
\newblock Faster r-cnn: Towards real-time object detection with region proposal
  networks.
\newblock In {\em Advances in Neural Information Processing Systems},
  volume~28, 2015.

\bibitem{ILSVRC15}
Olga Russakovsky, Jia Deng, Hao Su, Jonathan Krause, Sanjeev Satheesh, Sean Ma,
  Zhiheng Huang, Andrej Karpathy, Aditya Khosla, Michael Bernstein,
  Alexander~C. Berg, and Li Fei-Fei.
\newblock {ImageNet Large Scale Visual Recognition Challenge}.
\newblock {\em International Journal of Computer Vision (IJCV)},
  115(3):211--252, 2015.

\bibitem{sohn2020fixmatch}
Kihyuk Sohn, David Berthelot, Nicholas Carlini, Zizhao Zhang, Han Zhang,
  Colin~A Raffel, Ekin~Dogus Cubuk, Alexey Kurakin, and Chun-Liang Li.
\newblock Fixmatch: Simplifying semi-supervised learning with consistency and
  confidence.
\newblock {\em Advances in Neural Information Processing Systems}, 33, 2020.

\bibitem{tarvainen2017mean}
Antti Tarvainen and Harri Valpola.
\newblock Mean teachers are better role models: Weight-averaged consistency
  targets improve semi-supervised deep learning results.
\newblock In {\em Advances in neural information processing systems}, pages
  1195--1204, 2017.

\bibitem{tejankar2020isd}
Ajinkya Tejankar, Soroush~Abbasi Koohpayegani, Vipin Pillai, Paolo Favaro, and
  Hamed Pirsiavash.
\newblock Isd: Self-supervised learning by iterative similarity distillation,
  2020.

\bibitem{tian2019cmc}
Yonglong Tian, Dilip Krishnan, and Phillip Isola.
\newblock Contrastive multiview coding.
\newblock {\em arXiv preprint arXiv:1906.05849}, 2019.

\bibitem{tian2020makes}
Yonglong Tian, Chen Sun, Ben Poole, Dilip Krishnan, Cordelia Schmid, and
  Phillip Isola.
\newblock What makes for good views for contrastive learning?
\newblock In {\em Advances in Neural Information Processing Systems},
  volume~33, pages 6827--6839. Curran Associates, Inc., 2020.

\bibitem{oord2018representation}
Aaron van~den Oord, Yazhe Li, and Oriol Vinyals.
\newblock Representation learning with contrastive predictive coding, 2018.

\bibitem{wang2020invp}
Feng Wang, Huaping Liu, Di Guo, and Sun Fuchun.
\newblock Unsupervised representation learning by invariance propagation.
\newblock In {\em Advances in Neural Information Processing Systems},
  volume~33, pages 3510--3520. Curran Associates, Inc., 2020.

\bibitem{wei2020co2}
Chen Wei, Huiyu Wang, Wei Shen, and Alan Yuille.
\newblock Co2: Consistent contrast for unsupervised visual representation
  learning.
\newblock {\em arXiv preprint arXiv:2010.02217}, 2020.

\bibitem{wu2018unsupervised}
Zhirong Wu, Yuanjun Xiong, Stella~X Yu, and Dahua Lin.
\newblock Unsupervised feature learning via non-parametric instance
  discrimination.
\newblock In {\em Proceedings of the IEEE Conference on Computer Vision and
  Pattern Recognition}, pages 3733--3742, 2018.

\bibitem{sun397}
Jianxiong {Xiao}, James {Hays}, Krista~A. {Ehinger}, Aude {Oliva}, and Antonio
  {Torralba}.
\newblock Sun database: Large-scale scene recognition from abbey to zoo.
\newblock In {\em Computer Vision and Pattern Recognition}, 2010.

\bibitem{xie2016dec}
Junyuan Xie, Ross Girshick, and Ali Farhadi.
\newblock Unsupervised deep embedding for clustering analysis.
\newblock In {\em Proceedings of the 33rd International Conference on
  International Conference on Machine Learning - Volume 48}, ICML'16, page
  478–487. JMLR.org, 2016.

\bibitem{yangCVPR2016joint}
Jianwei Yang, Devi Parikh, and Dhruv Batra.
\newblock Joint unsupervised learning of deep representations and image
  clusters.
\newblock In {\em IEEE Conference on Computer Vision and Pattern Recognition
  (CVPR)}, 2016.

\bibitem{asano2020self}
Asano YM., Rupprecht C., and Vedaldi A.
\newblock Self-labelling via simultaneous clustering and representation
  learning.
\newblock In {\em International Conference on Learning Representations}, 2020.

\bibitem{zhang2016colorful}
Richard Zhang, Phillip Isola, and Alexei~A Efros.
\newblock Colorful image colorization.
\newblock In {\em European conference on computer vision}, pages 649--666.
  Springer, 2016.

\bibitem{zhang2017split}
Richard Zhang, Phillip Isola, and Alexei~A Efros.
\newblock Split-brain autoencoders: Unsupervised learning by cross-channel
  prediction.
\newblock In {\em Proceedings of the IEEE Conference on Computer Vision and
  Pattern Recognition}, pages 1058--1067, 2017.

\bibitem{zhuang2019local}
Chengxu Zhuang, Alex~Lin Zhai, and Daniel Yamins.
\newblock Local aggregation for unsupervised learning of visual embeddings.
\newblock In {\em Proceedings of the IEEE International Conference on Computer
  Vision}, pages 6002--6012, 2019.

\end{thebibliography}
}

\appendix
\onecolumn

\setcounter{figure}{0}
\renewcommand\thefigure{A\arabic{figure}}
\setcounter{table}{0}
\renewcommand\thetable{A\arabic{table}}

\section*{Appendix}

\begin{table*} [h]
    \begin{center}
    \scalebox{0.95}{
    \begin{tabular}{lrrrrrrr}
        \toprule
        Dataset & Classes & Train samples & Val samples & Test samples & Accuracy measure & Test provided \\
        \midrule
        Food101 [8] & 101 & 68175 & 7575 & 25250 & Top-1 accuracy & - \\
        CIFAR-10 [29] & 10 & 49500 & 500 & 10000 & Top-1 accuracy & - \\
        CIFAR-100 [29] & 100 & 45000 & 5000 & 10000 & Top-1 accuracy & -\\
        Sun397 (split 1) [48] & 397 & 15880 & 3970 & 19850 & Top-1 accuracy & - \\
        Cars [28] & 196 & 6509 & 1635 & 8041 & Top-1 accuracy & - \\
        Aircraft [30] & 100 & 5367 & 1300 & 3333 & Mean per-class accuracy & Yes \\
        DTD (split 1) [15] & 47 & 1880 & 1880 & 1880 & Top-1 accuracy & Yes \\
        Pets [36] & 37 & 2940 & 740 & 3669 & Mean per-class accuracy & - \\
        Caltech-101 [20] & 101 & 2550 & 510 & 6084 & Mean per-class accuracy & - \\
        Flowers [32] & 102 & 1020 & 1020 & 6149 & Mean per-class accuracy & Yes \\
        \bottomrule
    \end{tabular}
    }
    \caption{We list the sizes of train, val, and test splits of the transfer datasets. \textbf{Test split: } We use the provided test sets for Aircraft, DTD, and Flowers datasets. In case of Sun397, Cars, CIFAR-10, CIFAR-100, Food101, and Pets datasets, we use the provided val set as the hold-out test set. In case of Caltech-101, we use a random split of 30 images per category as the hold-out test set. \textbf{Val split: } We use the provided val sets for the datasets DTD and Flowers. For all other datasets, the val set is created by randomly sampling a subset of the train set. In order to be as close to BYOL [22] transfer setup as possible, we use the following val set splitting strategies for each dataset. Aircraft: 20\% samples/class. Caltech-101: 5 samples/class. Cars: 20\% samples/class. CIFAR-100: 50 samples/class. CIFAR-10: 50 samples/class. Food101: 75 samples/class. Pets: 20 samples/class. Sun397: 10 samples/class. }
    \label{tab:appendix_transfer_dset_details}
    \end{center}
\end{table*}

\begin{figure*}[!b]
\begin{center}
\end{center}
  \includegraphics[width=1.0\linewidth]{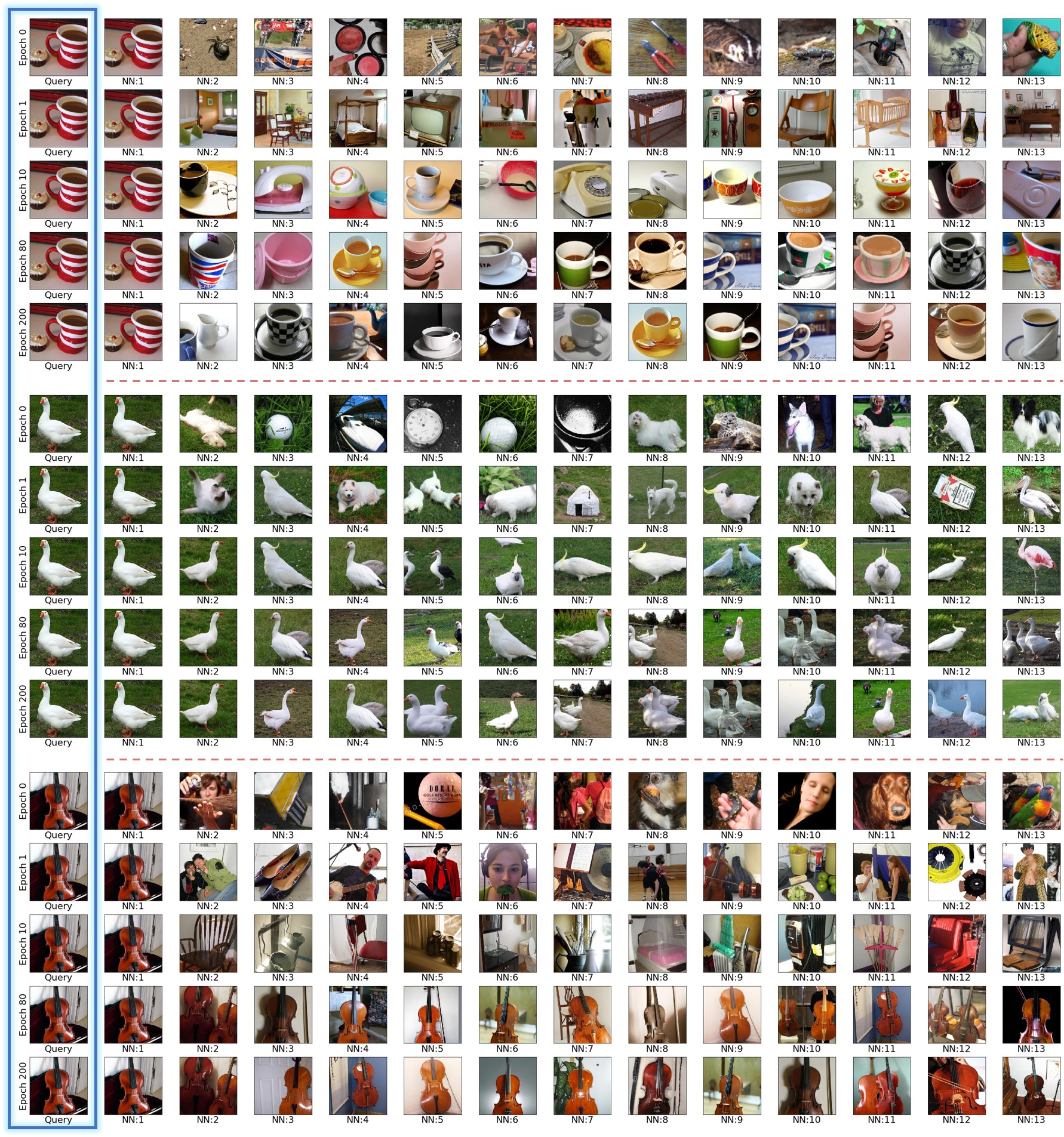}
  \caption{{\bf Nearest neighbors (NN) of the model at each epoch:} Similar to Figure 2.}
\label{fig:appendix_vis_NN_1}
\end{figure*}

\begin{figure*}[!b]
\begin{center}
\end{center}
  \includegraphics[width=1.0\linewidth]{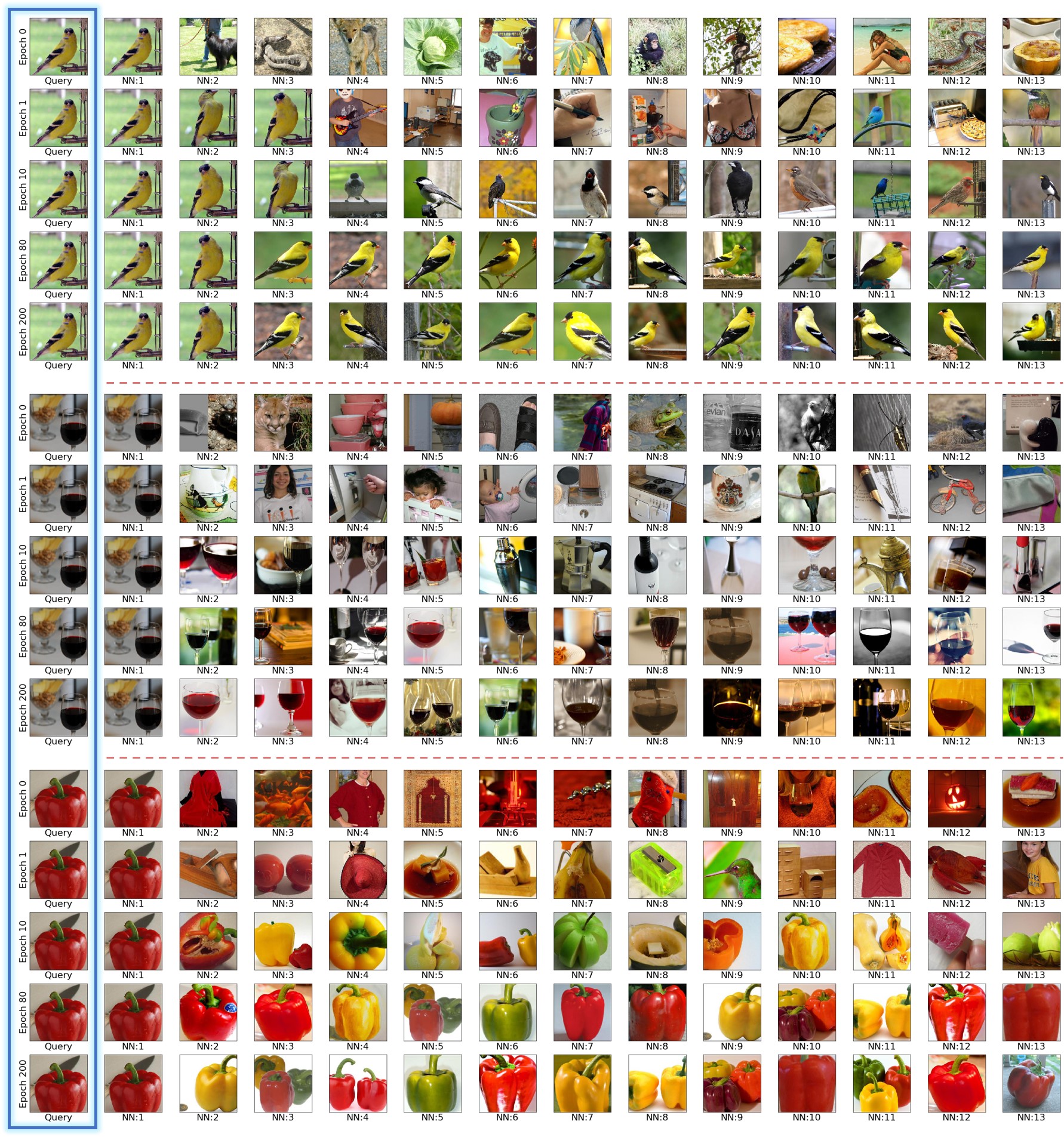}
  \caption{{\bf Nearest neighbors (NN) of the model at each epoch:} Similar to Figure 2.}
\label{fig:appendix_vis_NN_2}
\end{figure*}

\begin{figure*}[!b]
\begin{center}
  \includegraphics[width=0.85\linewidth]{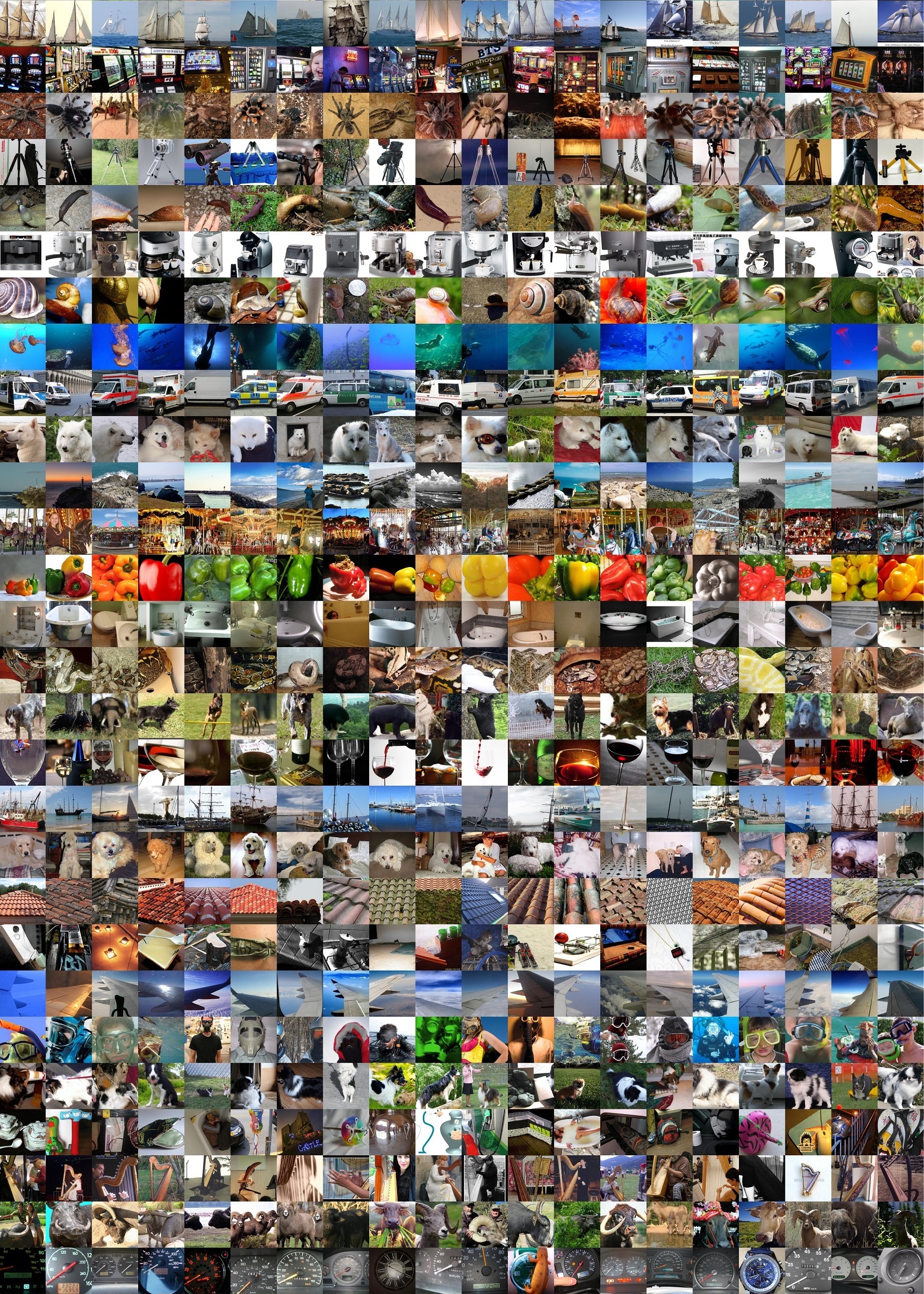}
  \caption{\textbf{Random Clusters:} We forward ImageNet training set through our ResNet-50 model and cluster them into 1000 clusters using k-means. We select 30 clusters randomly and show 20 randomly sampled images from each cluster without cherry-picking. Each row corresponds to a cluster. Note that semantically similar images are clustered together. 
  }
  \vspace{-.2in}
\label{fig:appendix_clusters}
\end{center}
\end{figure*}

\end{document}